\documentclass{article}
\usepackage{ijcai22}

\usepackage{multirow}
\usepackage[table,xcdraw]{xcolor}
\usepackage{graphicx}
\usepackage{wrapfig} 
\usepackage{subfigure}
\usepackage{times}
\usepackage{soul}
\usepackage{url}
\usepackage[hidelinks]{hyperref}
\usepackage[utf8]{inputenc}
\usepackage[small]{caption}
\usepackage{graphicx}
\usepackage{amsmath}
\usepackage{amsthm}
\usepackage{booktabs}
\usepackage{algorithm}
\usepackage{algorithmic}
\usepackage{amsfonts,amssymb,mathrsfs }
\usepackage{bm}
\usepackage{mathtools}
\usepackage{cuted}
\usepackage{lipsum}

\newtheorem{theorem}{Theorem}
\newtheorem{definition}{Definition}

\title{RoMFAC: A robust mean-field actor-critic reinforcement learning against adversarial perturbations on states}

\author{
	Ziyuan Zhou$^1$\and
	Guanjun Liu$^1$\footnote{Contact Author\\ The expanded version of this article is published in IEEE Transactions on Neural Networks and Learning Systems, with a DOI of 10.1109/TNNLS.2023.3278715.}\and
	\affiliations
	$^1$Department of Computer Sciences, Tongji University, Shanghai 201804, China\\
	\emails
	\{ziyuanzhou, liuguanjun\}@tongji.edu.cn,
}

\begin{document}

\maketitle

\begin{abstract}
  Multi-agent deep reinforcement learning makes optimal decisions dependent on system states observed by agents, but any uncertainty on the observations may mislead agents to take wrong actions. The Mean-Field Actor-Critic reinforcement learning (MFAC) is well-known in the multi-agent field since it can effectively handle a scalability problem. However, it is sensitive to state perturbations that can significantly degrade the team rewards. This work proposes a Robust Mean-field Actor-Critic reinforcement learning (RoMFAC) that has two innovations: 1) a new objective function of training actors, composed of a \emph{policy gradient function} that is related to the expected cumulative discount reward on sampled clean states and an \emph{action loss function} that represents the difference between actions taken on clean and adversarial states; and 2) a repetitive regularization of the action loss, ensuring the trained actors to obtain excellent performance. Furthermore, this work proposes a game model named a State-Adversarial Stochastic Game (SASG). Despite the Nash equilibrium of SASG may not exist, adversarial perturbations to states in the RoMFAC are proven to be defensible based on SASG. Experimental results show that RoMFAC is robust against adversarial perturbations while maintaining its competitive performance in environments without perturbations. 
\end{abstract}

\section{Introduction}
 Deep learning has achieved significant success in lots of fields such as computer vision and natural language processing.  However, deep neural networks are generally trained and tested by using independent and identically distributed data, and thus possibly make incorrect predictions when there are some small insignificant perturbations on samples \cite{goodfellow2014explaining,madry2018towards}. Deep learning has been combined with reinforcement learning  to train the control policy of an agent \cite{mnih2015human}. Some studies have shown that agents trained by deep reinforcement learning are also vulnerable to adversarial attacks, e.g., agents are likely to perform undesirable actions when their state space is perturbed \cite{9536399}. In fact, agents can frequently receive perturbed state observations because of sensor errors or malicious attacks, which can cause serious issues in many applications like autonomous driving, unmanned aerial vehicles and robotics. Therefore, robust deep reinforcement learning is very important in the single-/multi-agent filed.
 
 Many tasks require multiple agents to work together in a cooperative or competitive relationship rather than acting independently. The multi-agent reinforcement learning (MARL) is proposed to maximize team rewards. Many practical and effective MARL approaches  have been proposed, such as policy-based methods including MADDPG \cite{lowe2017multi}, MAAC \cite{iqbal2019actor} and G2ANet \cite{liu2020multi} and value-based methods including QMIX \cite{rashid2018qmix}, QPD \cite{yang2020q} and QPLEX \cite{wang2020qplex}. But they usually face a big challenge: the poor scalability, which significantly limits their applications in the real world.
 Mean-filed actor-critic method (MFAC) \cite{yang2018mean} applies the mean field theory to MARL and thus successfully improves the scalability of MARL with a large number of agents. However, this paper finds that MFAC is also sensitive to state perturbations which reduce its safety.

 Although there has been some progress in studies on adversary attacks and defenses of single-agent reinforcement learning algorithms, there are few related studies in the  multi-agent field. Compared with single-agent situations, multi-agent situations face additional challenges: 1) the total number of perturbed agents is unknown; and 2) perturbations on some agents can influence others. Facing these challenges, we propose a robust MFAC (RoMFAC) and 
our contributions are summarized  as follows:
\begin{itemize}
	\item We propose a novel objective function of training actors, which consists of a \emph{policy gradient function} that is related to the expected cumulative discount reward on sampled clean states and an \emph{action loss function} that represents the difference between actions taken on clean and adversarial states. We also design a repetitive regularization method for the action loss which ensures that the trained actors obtain a good performance not only on clean states but also on adversarial ones.
	\item We define the \emph{state-adversarial stochastic game} (SASG) by extending the objective function of RoMFAC to the stochastic game and study its basic properties which demonstrate that the proposed action loss function is convergent. Additionally, we prove that SASG dose not necessarily have the Nash equilibrium under the joint optimal adversarial perturbation but it can still defend against them. These theoretical results mean that our objective function can potentially be applied to some other reinforcement learning methods besides MFAC.
	\item We conduct experiments on two scenarios of MAgent \cite{zheng2018magent}. The experimental results show that our RoMFAC can well improve the robustness under white box attacks on states without degrading the performance on clean states.
\end{itemize}

\section{Related Works}
\paragraph{Adversarial Attacks on Single-agent DRL.}In classification tasks, the methods for generating and defending against adversarial examples have been extensively studied. Adversarial attacks and defenses for deep reinforcement learning have recently emerged. The adversarial attacks on DRL algorithms can be broadly divided into four categorizes \cite{9536399}: the state space with adversarial perturbations, the reward function with adversarial perturbations, the action space with adversarial perturbations and the model space with adversarial perturbations.
Huang \textit{et al.} \shortcite{huang2017adversarial} employ FGSM \cite{goodfellow2014explaining} to generate adversarial examples of input states, showing that adversarial attacks are also effective in the DRL policy network. To make the attack on DRL agents more stealthy and efficient, Sun \textit{et al.} \shortcite{sun2020stealthy} introduce two adversarial attack techniques: the critical point attack and the antagonist attack. This paper is about state perturbations.
\paragraph{Robust Training for Single-agent DRL.}Defense methods against attacks are broadly classified into six categories \cite{9536399}: adversarial training, defensive distillation, robust learning, adversarial detection, benchmarking \& watermarking and game theoretic approach. 
Zhang \textit{et al.} \shortcite{zhang2020robust} propose state-adversarial Markov decision process (SA-MDP), which provides a theoretical foundation for robust single-agent reinforcement learning. They develop the principle of policy regularization that can possibly be applied to many DRL algorithms. Based on SA-MDP, an alternate training framework with learned adversaries was proposed \cite{zhang2021robust}. Oikarinen \textit{et al.} \shortcite{oikarinen2020robust} propose the RADIAL-RL method, which can improve the robustness of DRL agents under the $\ell_p$ norm boundary against attacks, with lower computational complexity. This paper focuses on robust learning and expands the theoretical results and policy regularization in SA-MDP to multi-agent DRL.
\paragraph{Adversarial Attacks and Defenses for Multi-agent DRL.}Motivated by single-agent deep reinforcement learning \cite{mnih2015human}, multi-agent reinforcement learning has changed from the tabular method to the deep learning recently. 
However, there are few studies on adversarial attacks and robust training in multi-agent DRL. Lin \textit{et al.} \shortcite{lin2020robustness} first propose the method of adversarial examples generation in MARL, but do not provide a robust defense method. Li \textit{et al.} \shortcite{li2019robust} propose the M3DDPG, which is an extension of MADDPG that makes policies of agents generalizing even if the opponent's policies change. They also present a robust optimization method, which effectively solves the problem of high complexity of minmax calculation in continuous action space. However, they lack the defense against state perturbations that is exactly the purpose of our work.
\section{Preliminary}
\subsection{Stochastic Game}
Stochastic game (SG) \cite{shapley1953stochastic} is a game with multiple agents (or players) and states, defined as a tuple $\left<\mathcal S, \mathcal A^1,\ldots,\mathcal A^N,R^1,\ldots,R^N,p,\gamma\right> $ where $ \mathcal S $ is the state space, $N$ is the number of agents, $\mathcal A^j$ is the action space of agent $j$, $ R^j: \mathcal S\times\mathcal A^1\times \cdots \times \mathcal A^N\times\mathcal S\rightarrow\mathbb R $ is the reward function of agent $j$, $ p: \mathcal S\times\mathcal A^1\times \cdots \times \mathcal A^N\times\mathcal S\rightarrow\left[0,1\right] $ is the state transition probability function which refers to the probability distributions of the next states under the current state and the joint action, and $\gamma\in\left[0,1\right] $ is the discount factor. The immediate reward $ R^j\left(s,\bm a,s'\right) $ represents the reward obtained by agent $j$ in state $s’$ after taking the joint action $ \bm {a}\buildrel\Delta\over =\left(a^1,\ldots,a^N\right)$ in state $s$.

For an $n$-player stochastic game, there is at least one Nash equilibrium \cite{1964Equilibrium}, which can be defined as the joint policy $ \bm \pi_*\buildrel\Delta\over =\left(\pi^1_*,\ldots,\pi^N_*\right) $ such that $\forall s\in\mathcal S$: 
\begin{equation}
	\begin{aligned}
		V^j_{\bm \pi_*}\left(s\right)&=V^j\left(s,\pi^1_*,\ldots,\pi_*^j,\ldots,\pi^N_*\right)\\
		&\geq V^j\left(s,\pi^1_*,\ldots,\pi^j,\ldots,\pi^N_*\right)
	\end{aligned}
\end{equation}
where $\bm \pi_*\left(\cdot|s\right)=\prod_{j=1}^N\pi_*^j\left(\cdot|s\right)$ is the probability distribution of the joint action $\bm a$ at state $s$ under the Nash equilibrium and $\pi^j$ is an arbitrary valid policy of agent $j$. $V^j_{\bm \pi_*}\left(s\right)$ is the value function of agent $j$ under state $s$ and the Nash equilibrium at time $t$ and calculated through  expected cumulative discount reward of agent $j$:
\begin{equation}\label{v}
	V^j_{\bm \pi_*}\left(s\right)=\mathbb E_{\bm \pi_*,p}\left(\sum_{k=0}^\infty \gamma^kR^j_{t+k+1}|s_t=s\right) 
\end{equation}
where $R^j_{t}$ denotes the reward of agent $j$ at the time $t$. The action-value $Q^j_{\bm \pi_*}\left(s,\bm a\right)$ is defined as the expected cumulative discount reward of agent $j$ given a state $s$ and a joint action $\bm a$ of all agents under the Nash equilibrium: 
\begin{equation}\label{q}
	\begin{aligned}
		Q^j_{\bm \pi_*}\left(s,\bm a\right)&= \mathbb E_{\bm \pi_*,p}\left(\sum_{k=0}^\infty \gamma^kR^j_{t+k+1}|s_t=s,\bm a_t=\bm a\right)\\
		&=\sum_{s'\in\mathcal S}p\left(s'|s,\bm a\right)\left(R^j\left(s,\bm a,s'\right)+\gamma V^j_{\bm \pi_*}\left(s'\right)\right).
	\end{aligned}
\end{equation}
According to Eqs.~(\ref{v}) and (\ref{q}), the value function can also be formulated as 
\begin{equation}
	V^j_{\bm \pi_*}\left(s\right)=\mathbb E_{\bm a \sim\bm \pi_*}\left(Q^j_{\bm \pi_*}\left(s,\bm a\right)\right).
\end{equation}

\subsection{Mean-Field Actor-Critic Reinforcement Learning}
Mean-field actor-critic reinforcement learning (MFAC) \cite{yang2018mean} uses the mean-field theory to transform the interaction of multiple agents into the interaction between two agents, which makes large-scale multi-agent reinforcement learning become possible. In MFAC, $Q^j\left(s,\bm a\right)$ is decomposed into $$ 
Q^j(s,\bm a)=\frac{1}{|\mathcal N(j)|}\sum_{k\in\mathcal N(j)}{Q^j(s,a^j,a^k)}$$ through local interactions, where $\mathcal N(j)$ is the set of neighbors of agent $j$. They prove that $$Q^j\left(s,\bm a\right)\approx Q^j\left(s,a^j,\bar a^j \right)$$
where the mean action $\bar a^j$ of all neighbors of agent $j$ can be represented as an empirical distribution of the actions taken by these neighbors and obtained by calculating the average of $a^k$, while $a^k$ is sampled from policy $\pi^k$ which is calculated by a neural network via the previous average action $\bar a_-^k $ of agent $j$’s neighbors: $${\bar a^j=\frac{1}{|\mathcal N(j)|}\sum_{k,{a^k\sim\pi^k\left(\cdot|s,\bar a^k_-\right)}} a^k}.$$
Note that each $a^k$ is a one-hot coding. Then the policy $\pi^j$ is changed according to the current $s$ and $\bar a^j$.

The mean field Q-function at time $t$ can be updated according to the following recursive form:
\begin{equation}\label{mfq}
	Q_{\phi^j}\left(s,a^j,\bar a^j  \right)_{t+1} \!\! =\!\left( \! 1\!-\!\alpha\right)\!Q_{\phi^j}\!\!\left(s,a^j,\bar a^j\right)_t \!+\alpha\!\left( \!R^j \!\!+\!\!\gamma V^j\!\! \left(s'\right)_t\right)
\end{equation}
where $\alpha$ is the learning rate and $\phi^j$ is the parameters of the critic of agent $j$. The mean-filed value function at time $t$ can be calculated as
\begin{equation}\label{mfv}
	V^j\left(s'\right)_t \! = \! \sum_{a^j} \! \pi_{\theta^j}\!\!\left(a^j|s',\bar a^j\right)_t\!\mathbb E_{\bm a^{-j}\sim \bm \pi_{\theta^{-j}}}\!\!\left(Q_{\phi^j}\!\!\left(s',a^j,\bar a^j\right)_t\right)
\end{equation}
where $\theta^j$ is the parameters of the actor of agent $j$, $\bm a^{-j}$ is the joint action of all agents expect agent $j$ and $\bm \pi_{\theta^{-j}}$ is the joint policy of all agents expect agent $j$. 

MFAC is an on-policy actor-critic method where the critic is trained by minimizing the loss function: 
\begin{equation}\label{closs}
	\mathcal L_{crt}(\phi^j)=(y^j-Q_{\phi^j}(s,a^j,\bar a^j))^2
\end{equation}
and the actor $\pi_{\theta^j}$ is trained by sampling policy gradients:
\begin{equation}\label{aloss}
	\begin{aligned}
		\nabla_{\theta^j}\mathcal J(\theta^j)&\buildrel\Delta\over =\nabla_{\theta^j} V_{\pi_{\theta^j}}\left(s\right) \\
		&\approx \nabla_{\theta^j}\left(\log\pi_{\theta^j}(s)\right)Q_{\phi_-^j}(s',a_-^j,\bar a^j_-)|_{a_-^j=\pi_{\theta_-^j}(s)}
	\end{aligned}
\end{equation}
where $\phi_-^j$ and $\theta_-^j$ are parameters of target networks of agent $j$. \cite{konda2000actor} provides a derivation of Eq.~(\ref{aloss}). During the training  process, $\phi^j$ and $\theta^j$ are alternately updated until convergence is achieved. Since the observed states are perturbed, this paper proposes a new objective function to train an actor so that it can defend against state perturbations.
\section{Methodology}
\subsection{Robust Mean-Field Actor-Critic Method} \label{4.1}
In this section, we introduce a novel framework for improving the robustness of MFAC on state perturbations. Our framework mainly contains the following innovative components:

\subsubsection{Action Loss Function}
In order to learn a robust policy, we propose a novel method to update the policy network. In the worst case, states of every agent are all attacked and thus we should optimize the expected cumulative discount reward corresponding to adversarial states. According to Eq.~(\ref{mfv}), the learning objective of our RoMFAC is to maximize the expected cumulative discount reward in the worst-case, i.e., 
\begin{equation}\label{minmax}
	\begin{aligned}
		& \;\;\;\;\max_{\theta^j} V^j\left(s\right)\\
		&= \max_{\theta^j} \min_{\hat {s}^j} \sum_{a^j}\!\!\pi_{\theta^j}\!\!\left(a^j|\hat s^j,\bar a^j \right)\!\mathbb E_{\bm{a}^{\!-j}\sim\bm\pi_{\theta^{-j}}}\!\!\left(Q_{\phi^j}\!\!\left(s,a^j,\bar a^j\!\right)\!\right)
	\end{aligned}
\end{equation}
where $\hat{s}^j$ is the adversarial state of agent $j$ calculated by Eq.~(\ref{pgd}) and the goal is to minimize the expected cumulative discount reward of agent $j$. Since critics are only used to guide the update of actors on which behaviors of agents depend, here we only consider the robustness of actors. The minimized part of Eq.~(\ref{minmax}) can be solved by maximizing the loss between actions taken on clean and adversarial states, and the loss is called \emph{action loss} in this paper, i.e., 
\begin{equation} \label{advloss}
	\mathcal L_{act} \left(\theta^j\right)=\max \limits_{\hat{s}^j \in \mathcal{B}^j} \left\{L\left(\theta^j, \hat{s}^j, z^j\right)| \hat{s}^j=s+\delta^j\right\}
\end{equation}
where $s$ is the clean state, $\delta^j$ is the adversarial perturbation of agent $j$ generated according to state $s$, and $\mathcal{B}^j$ is the set of adversarial states of agent $j$. We can label the action with the highest probability $z^j=\arg \max\limits_{a^j} \pi_{\theta^j}\left(a^j|s,\bar{a}^j\right)$, because the actor of agent $j$ outputs the probability distribution of actions.
$L$ is the cross-entropy loss function of actions taken on clean and adversarial states, i.e.,
$$
	L\left(\theta^j, \hat s^j, z^j\right)=-\sum_{a^j} \left[z^j = a^j\right] \log\left({\pi_{\theta^j}}\left(a^j|\hat s^j,\bar{a}^j\right)\right)
$$
where $\left[z^j = a^j\right]$ is the Iverson bracket whose value is $1$ if the statement $z^j=a^j$ is true, and $0$ otherwise.

To solve the maximization problem, we use the PGD method \cite{madry2018towards} to generate the adversarial perturbation $\delta^j$ for the policy network of agent $j$. The PGD uses multi-step gradient ascent:
\begin{equation} \label{pgd}
	\hat s_{u+1}^j=\mbox{clip}\left(\hat s_{u}^j +  \beta \mbox{sgn}\left(\nabla_{\hat s_{u}^j}  L\left(\theta,\hat s_{u}^j,z^j\right) \right)\right)|_{\hat s^j_0=s}
\end{equation}
where $\beta$ is the step-size and $\hat s_u^j$ represents the adversarial state in the $u$-th step initialized by $s$.
The outer-loop learning objective is to minimize the difference of actions taken on adversarial and clean states. The actor $\pi_{\theta^j}$ is trained by minimizing
\begin{equation} \label{actor_loss}
	-\nabla_{\theta^j}\mathcal J(\theta^j)+\mu\mathcal L_{act}(\theta^j)
\end{equation}
where $\mu$ is a weight factor governing the trade-off between the two parts. 
\linespread{0.5}
\begin{algorithm}[t]
	\caption{RoMFAC}
	\label{alg:algorithm}
	\begin{algorithmic}[1] 
		\STATE Initialize $Q_{\phi^j}$, $Q_{\phi^j_-}$, $\pi_{\theta^j}$, $\pi_{\theta^j_-}$ and $\bar a^j$, $ \forall j \!\in \! \{1,\dots,N\}$
		\FOR{$m=1,2,\dots,M_{norm}+cM_{adv}$}
		\STATE For each agent $j$, sample action $a^j=\pi_{\theta^j}(s)$ and compute the new mean action $\bar {\bm a}=[\bar a^1,\dots,\bar a^N]$;
		\STATE Take the joint action $\bm {a}=[a^1,\dots,a^N]$ and observe the reward $\bm {R}=[R^1,\dots, R^N]$ and the next state $s'$;
		\STATE Store $\left<s,\bm a, \bm R,s',\bar {\bm a}\right>$ in replay buffer $\mathcal D$;
		\FOR{$j=1$ to $N$}
		\STATE Sample a minibatch of $K$ experiments $\left<s_i,\bm a_i, \bm R_i,{s'}_i,\bar {\bm a}_i\right>|_{i=1,\dots,K}$ from $\mathcal D$;
		\STATE Set $ y^j=R^j+\gamma V_{\phi _-^j}(s')$;
		\STATE Update the critic based on Eq.~(\ref{closs}); 
		\STATE Generate adversarial perturbation $\delta^j$ and get adversarial state $\hat{s}^j_i=s_i+\delta^j$ based on Eq.~(\ref{pgd});
		\STATE Compute the action loss $\mathcal L_{act}^i(\theta^j,\hat s_i^j)$;
		\STATE Update the actor based on Eqs.~(\ref{actor_loss}) and (\ref{mu}); 
		\ENDFOR
		\STATE Update the parameters of the target networks for each agent $j$ with learning rates $\tau_\phi$ and $\tau_\theta$:
		
		\centerline{$\phi_-^j\gets\tau_{\phi}\phi^j+(1-\tau_\phi)\phi_-^j$}
		
		\centerline{$\theta_-^j\gets\tau_\theta\theta^j+(1-\tau_\theta)\theta^j_-$}
		\ENDFOR
	\end{algorithmic}
\end{algorithm}

\subsubsection{Repetitive Regularization of the Action Loss}

\begin{figure}[t]
	\centering
 	\includegraphics[scale=0.71]{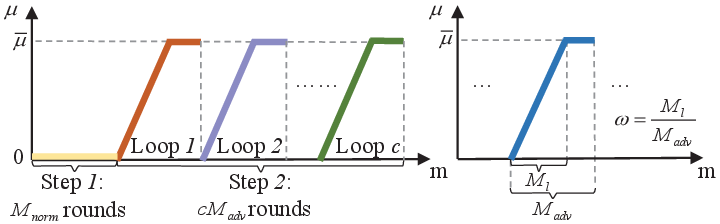}
	\caption{Repetitive change of $\mu$}
	\label{fmu}
\end{figure}
For the weight factor $\mu$ of the action loss, if it is too large, there may be vanishing and exploding gradient and thus the training is unstable. On the other hand, if it is too small, the action loss will not work. 
Regularizing the loss related to adversarial perturbations is often used in many robust single-agent reinforcement learning \cite{zhang2020robust,zhang2021robust}, but they usually use the grid search method to produce a fixed value for the weight factor $\mu$, and the perturbation bound $\epsilon$ gradually increases to a given value in the whole training process. Our experiments indicate that if the input data of a multi-agent environment is not high-dimensional image data, training results obtained by this way are not ideal. 

To solve this problem, we propose a repetitive regularization method for our action loss: 
\begin{itemize}
	\item[]Step 1: We train a network until it is stable with $\mu=0$;
	\item[]Step 2: We continually train it by $c$ loops. In every loop, $\mu$ increases linearly from 0 to a given upper bound $\overline{\mu}$.
\end{itemize}
During the whole training process, the perturbation bound $\epsilon$ is a fixed value. In addition, $c$ and $\overline{\mu}$ are two hyper-parameters.
Figure \ref{fmu} shows the idea of our repetitive regularization method. In fact, $\mu$ can be calculated by the following formula for the $m$-th round:
\begin{equation} \label{mu}
	\mu\left(m \right)=\frac{\min\! \left\{\! \max \!\left\{m-M_{norm},0\right\}\!\!\!\!\! \mod M_{adv}, \omega M_{adv}\right\}}{\omega M_{adv}}\overline{\mu}
\end{equation}
where $M_{norm}$ is the number of training rounds in Step 1 and $M_{adv}$ is the number of training rounds in every loop. This repetitive change can make agents explore more new positive behaviors while simultaneously increasing the robustness against adversarial states. 

Our RoMFAC is presented in Algorithm \ref{alg:algorithm}. In the next section, we demonstrate that  minimizing the learning objective of Eq.~(\ref{minmax}) is convergent.
\subsection{State-Adversarial Stochastic Game}
\begin{figure}[t]
	\centering
	\includegraphics[scale=0.45]{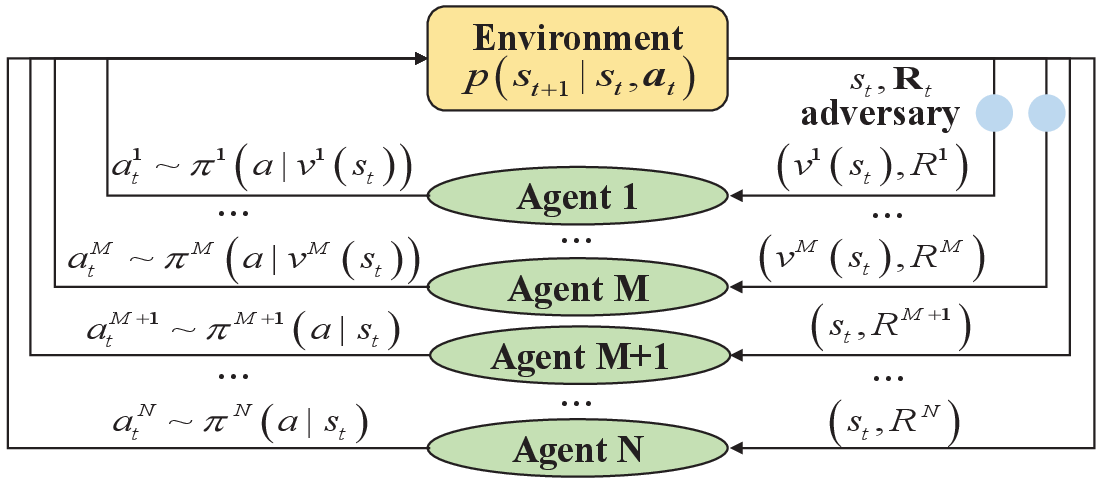} 
	\caption{The illustration of State-Adversarial Stochastic Game.} 
	\label{SA-SG} 
\end{figure}
In this section, we define a class of games: SASG (state-adversarial stochastic game) to which the objective function in Section \ref{4.1} is applied. SASG allows adversarial perturbations, and we prove that adversarial perturbations can be defended in theory.
\begin{definition}[SASG]
	An SASG can be defined as a tuple $\left<\mathcal S, \mathcal A^1,\dots,\mathcal A^N,\mathcal B^1,\dots, \mathcal B^M, R^1,\dots,R^N,p,\gamma \right>$. $\mathcal B^j$ is the set of adversarial states of agent $j$, and $M$ is the number of attacked agents and $M \le N$. 
\end{definition}
We define the adversarial perturbation $v^j(s)$ of agent $j$ as a deterministic function, that is, it is only dependent on the current state $s$ and does not change over time, $v^j:\mathcal S \rightarrow \mathcal B^j$. As shown in Figure \ref{SA-SG}, $v^j(s)$ only perturbs the state of agent $j$, while the environment itself keeps unchanged. 
The value and action-value function of SASG are similar to SG:
$$
\tilde V_{\bm \pi \circ \bm v}^j\left(s\right)=\mathbb E_{\bm \pi \circ \bm v}\left(\sum_{k=0}^\infty \gamma^kR^j_{t+k+1}|s_t=s\right),
$$
$$
\tilde Q_{\bm \pi \circ \bm v}^j\left(s,\bm a\right)=\mathbb E_{\bm \pi \circ \bm v}\left(\sum_{k=0}^\infty \gamma^kR^j_{t+k+1}|s_t=s,\bm a_t= \bm a\right)
$$
where $\bm v\buildrel\Delta\over= \left(v^1,\cdots,v^M\right)$ denotes the joint adversarial perturbation and $\bm \pi \circ \bm v$ denotes the joint policy under the joint adversarial perturbation: $\bm \pi \circ \bm v \buildrel\Delta\over= \bm \pi\left(\cdot|s , {\bm v(s)}\right) = \prod_{j=1}^M{\pi^j\left(\cdot|v^j \left(s \right) \right)}\prod_{j=M+1}^N{\pi^j\left(\cdot|s\right)}$.

The proofs of the following conclusions of SASG are put into Appendices A--D of our supplementary file.
\begin{theorem}[Bellman equations of fixed $\bm \pi$ and $\bm v$] \label{t1}
	Given the joint policy $\bm \pi: \mathcal S \rightarrow \mbox{PD} \left(\mathcal A^1\times \cdots \times \mathcal A^N \right)$ and $\bm v:\mathcal S \rightarrow \mathcal B^1\times \cdots \times \mathcal B^M $, we have
	\begin{equation}
		\begin{aligned}
			\tilde{V}^j_{\bm \pi \circ \bm v}(s)=\sum_{\bm a  \in \mathcal A^1  \times \cdots  \times  \mathcal A^N} \bm \pi\left(\bm a|s ,\bm v(s)\right)\sum\limits_{s' \in \mathcal S}\left( p\left(s'|s,\bm a\right) 
			\notag\right.
			\\
			\phantom{=\;\;}
			\left.\left(R^j\left(s,\bm a,s'\right)+\gamma\tilde{V}^j_{\bm \pi \circ \bm v}(s')\right)\right),
		\end{aligned}
	\end{equation}%
	$$
	\tilde Q_{{\bm{\pi }} \circ {\bm {v}}}^j\left( s \right) =\sum\limits_{s' \in \mathcal S} p\left(s'|s,\bm a\right)\left( {R^j} \left( {s,{\bm{a}},s'} \right) + \gamma {\tilde V_{{\bm{\pi }} \circ {\bm {v}}}^j\left( {s'} \right)}\right).
	$$
\end{theorem}
The goal of the joint optimal adversarial perturbation is to minimize expected cumulative discount reward of every attacked agent, and hence the value function and action-value function can be written as
$$
\tilde V_{{\bm{\pi }} \circ {{\bm{v}}_*}}^j\left( s \right) = \min \tilde V_{{\bm{\pi }} \circ\left( {v^j,\bm{v}^{-j}_*}\right)}^j\left( s \right),$$
$$\tilde Q_{{\bm{\pi }} \circ {{\bm{v}}_*}}^j\left( s, \bm{a} \right) = \min \tilde Q_{{\bm{\pi }} \circ\left( {v^j,\bm{v}_*^{-j}}\right)}^j\left( s,\bm{a} \right)
$$
where $\bm v_*\buildrel\Delta\over= \left(v^1_*,\cdots,v^M_*\right)$ is the joint optimal adversarial perturbation, $v^j$ is an arbitrary valid adversarial perturbation and $\bm v_*^{-j}\buildrel\Delta\over= \left(v_*^1,\cdots,v_*^{j-1},v_*^{j+1},\cdots,v_*^M\right)$. Obviously, the minimized part of Eq.~ (\ref{minmax}) is a special case of Theorem \ref{t1}.
\begin{theorem}[Bellman contraction of agent $j$ for the joint optimal adversarial perturbation]\label{t2}
	Define Bellman operator $\mathscr{L}^j: \mathbb{R}\rightarrow \mathbb{R}$,
	\begin{equation}
			\begin{aligned}
				\left( {\mathscr L^j{{\tilde V}^j}} \right)\!\left( s \right)\! = \!\!\! \mathop {\min }\limits_{ v^j(s)  \in \mathcal B^j}   \!\!  \sum\limits_{{\bm{a}} \in {\mathcal{{\cal A}}^{_1}} \times  \cdot  \cdot  \cdot  \times {\mathcal{{\cal A}}^{_N}}} \!\!\!\!\!\!\!\!\!\! \left({\bm{\pi }}\left( {\bm a|s, v^j(s) , \bm v^{-j}_*\left( s \right)} \right) 
				\notag\right.
				\\
				\phantom{=\;\;}
				\left.\sum\limits_{s' \in S} {p\left( {s'|s,{\bm{a}}} \right)\left( {{R^j}\left( {s,{\bm{a}},s'} \right) + \gamma {{\tilde V}^j}\left( {s'} \right)} \right)}\right).
			\end{aligned}
	\end{equation}%
	Then, the Bellman equation for the joint optimal adversarial perturbation $\bm v_*$ is $\tilde V_{{\bm{\pi }} \circ {{\bm{v}}_*}}^j = \mathscr{L}^j\tilde V_{{\bm{\pi }} \circ {{\bm{v}}_*}}^j$. Furthermore, $\mathscr{L}^j$ is a contraction that converges to $\tilde V_{{\bm{\pi }} \circ {{\bm{v}}_*}}^j$.
\end{theorem}
	Theorem \ref{t2} indicates that $\mathscr{L}^j$ converges to a unique fixed point, that is, the joint optimal adversarial perturbation $\bm v_*\left(s\right)$ is unique. Consequently, the proposed action loss (i.e., Eq.~(\ref{advloss})) is convergent since there is a unique solution to $\hat s^j$ that leads to the worst case of Eq.~(\ref{minmax}).  
\begin{theorem}[]\label{t3}
	Under the joint optimal adversarial perturbation $\bm v_*$, the Nash equilibrium of SASG may not always exist.
\end{theorem}
\begin{theorem}[]\label{t4}
	Given the joint policy $\bm \pi$, under the joint optimal adversarial perturbation $\bm v_*$, we have
		\begin{equation}
			\begin{aligned}
			&\;\;\;\;\mathop {\max }\limits_{s \in \mathcal{{\cal S}}} \left\{ {\tilde V_{\bm{\pi }\circ\bm v_*^{-j}\left(s\right)}^j\left( s \right) \!- \!\tilde V_{{\bm{\pi }} \circ {{\bm{v}}_*}}^j \! \left( s \right)} \right\}  \\
			&\le{\zeta}\mathop {\max }\limits_{s \in \mathcal{{\cal S}}}\mathop {\max }\limits_{ {\hat s^j} \in \mathcal B^j }\!{\text{D}_{\text{TV}}}\left( {{\bm{\pi }}\left( {\cdot|s , \bm {\hat s}^{-j} } \right),{\bm{\pi }}\left( {\cdot|s , \hat s^j , \bm {\hat s}^{-j} } \right)} \right)
		\end{aligned}
		\end{equation}
	where $ {\text{D}_{\text{TV}}}$ is the total variation, $\bm {\hat s}^{-j}$ is a group of adversarial states of all attacked agents expect the agent $j$, i.e., $\bm {\hat s}^{-j} =\bm v_*^{-j}\left(s\right)$, $\hat s^j$ is an arbitrary valid adversarial state of agent $j$, 
	and $\zeta\buildrel \Delta \over = 2\left( {1 + \frac{\gamma }{{{{\left( {1 - \gamma } \right)}^2}}}} \right)\mathop {\max }\limits_{s,{\bm{a}},s'} \left| {{R^j}\left( {s,{\bm{a}},s'} \right)} \right|$ is a constant independent of $\pi^j$.
\end{theorem}
Theorem \ref{t4} indicates that when there are adversarial states, the intervention of the value function is small as long as the difference between action distributions is small. Therefore, we can train robust policies, even if there is possibly no Nash equilibrium under the joint optimal adversarial perturbation as shown in Theorem \ref{t3}. These conclusions also mean that our robust method can be applied to some other multi-agent reinforcement learning.
\section{Experiments}
We demonstrate the superiority of RoMFAC in improving model robustness against adversarial perturbations. 

\subsection{Environments}
We use two scenarios of MAgent \cite{zheng2018magent} which can support hundreds of agents for our experiments. 
\paragraph{Battle.}This is a cooperative and competitive scenario in which two groups of agents, A and B, interact. Each group of agents works together as a team to eliminate all opponent agents. There are $128$ agents in total and $64$ ones in each group. We use the default reward settings: $0.005$ per step, $0.2$, $5$ for attacking or killing an opponent agent, $-0.1$ for attacking an empty grid, and $-0.1$ for being attacked or killed.
\paragraph{Pursuit.} This is a scenario of local cooperation. There are $32$ predators and $64$ prey. Similarly, we use the default reward setting: the predator receives $+1$ for attacking the prey, while the prey receives $-0.1$ for being attacked.
\begin{figure}[t]
	\centering
	\subfigure[MFAC in battle.]{
		\begin{minipage}[h]{\linewidth}
			\centering
			\includegraphics[scale=0.35]{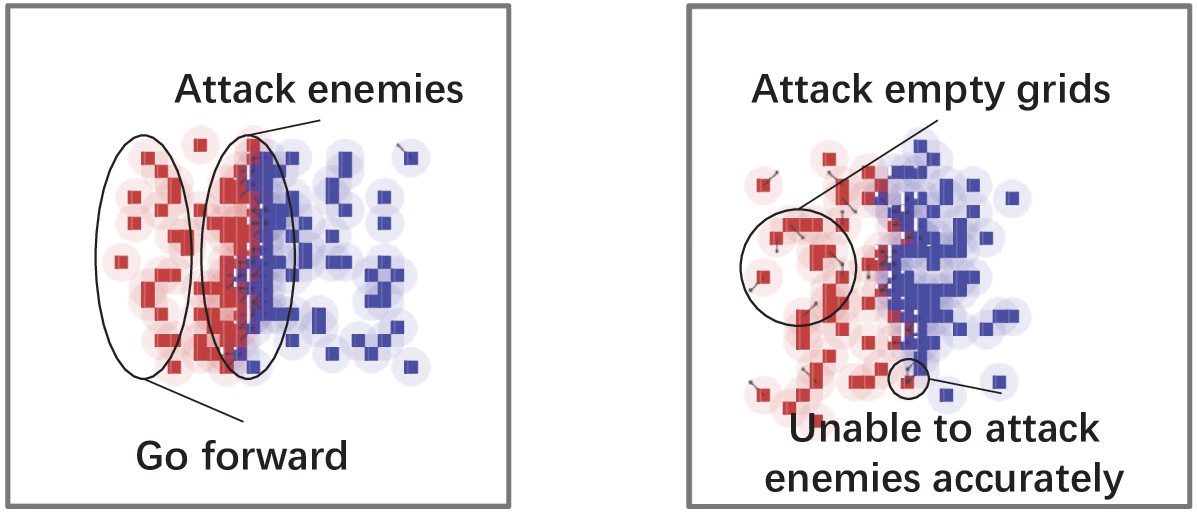}
			\label{11}
		\end{minipage}%
	}%

	\subfigure[MFAC in pursuit.]{
		\begin{minipage}[h]{\linewidth}
			\centering
			\includegraphics[scale=0.35]{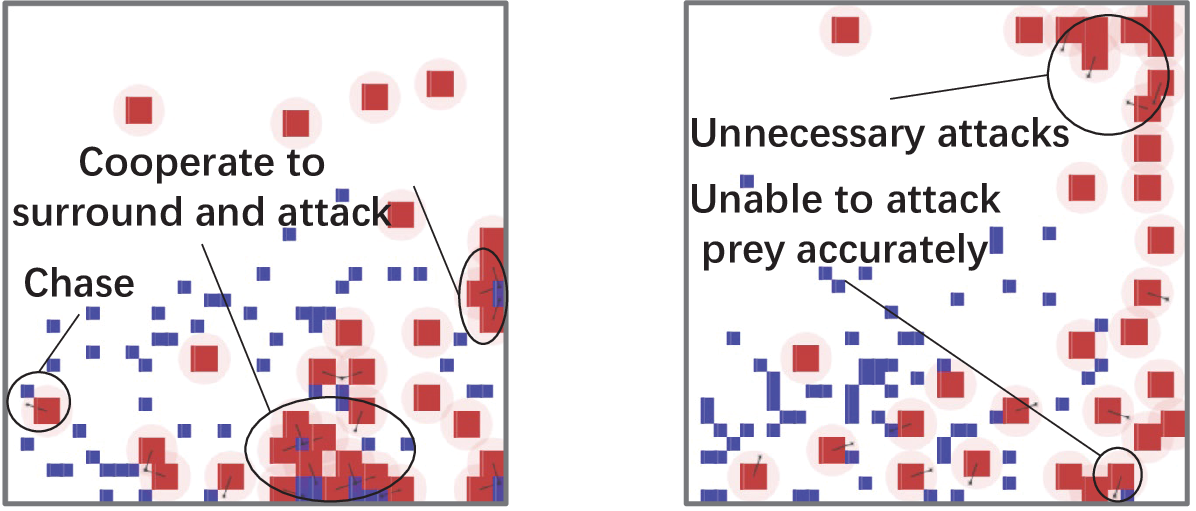}
			\label{22}
		\end{minipage}%
	}%
	\caption{Illustration of representative behaviors of MFAC agents in battle and pursuit scenarios. The left sides of (a) and (b) are behaviors of MFAC agents under clean states, and the right sides are behaviors under adversarial states.}
	\label{figscenarios}
\end{figure}

\subsection{Evaluation with Adversarial States}

\begin{table}[t]
	\centering
	\scalebox{0.675}{
		\begin{tabular}{l|l|rrr|r}
		\toprule
		Nos. of &      & \multicolumn{3}{r|}{\cellcolor[HTML]{FFFFFF}Battle}                  &\multicolumn{1}{r} {Pursuit}                     \\ 
		attacked                                 & Methods     & Wining    & Average          & Average      & Average         \\ 

		agents &     & rate   & kill            & total reward      & total reward        \\
		\midrule
		\cellcolor[HTML]{FFFFFF}                     & MFAC        & 0.66          & 61.80$\pm$3.36          & 294.21$\pm$19.13          & 3674.04$\pm$498.83          \\ 
		\cellcolor[HTML]{FFFFFF}                     & SA-MFAC     & 0.52          & 59.28$\pm$5.96          & 294.53$\pm$28.88          & 3036.41$\pm$427.37       \\ 
		\cellcolor[HTML]{FFFFFF}                     & SA-MFAC$^3$ & 0.30         & 57.20$\pm$5.28         & 282.71$\pm$26.99        & 3619.32$\pm$442.18     \\
		\cellcolor[HTML]{FFFFFF}                     & RoMFAC$^1$   & 0.52          & 59.76$\pm$5.20          & 297.65$\pm$27.13          & 3282.41$\pm$472.68          \\ 
		\multirow{-14}{*}{\cellcolor[HTML]{FFFFFF}0}  & RoMFAC & \textbf{1.00} & \textbf{63.98$\pm$0.14} & \textbf{320.29$\pm$8.56}  & \textbf{3844.66$\pm$462.89} \\ 
		\midrule  
		\cellcolor[HTML]{FFFFFF}                     & MFAC        & 0.48          & 59.24$\pm$5.64          & 279.73$\pm$29.46          & 3012.28$\pm$377.84          \\ 
		\cellcolor[HTML]{FFFFFF}                     & SA-MFAC     & 0.42          & 58.60$\pm$5.47          & 287.78$\pm$29.59          & 3031.31$\pm$454.87      \\
		\cellcolor[HTML]{FFFFFF}                     & SA-MFAC$^3$ &0.44          & 57.52$\pm$5.42         & 279.30$\pm$25.53         & 3440.18$\pm$450.34     \\ 
		\cellcolor[HTML]{FFFFFF}                     & RoMFAC$^1$    & 0.52          & 60.34$\pm$4.38          & 298.44$\pm$22.80          & 3453.61$\pm$446.54          \\ 
		\multirow{-14}{*}{\cellcolor[HTML]{FFFFFF}8}  & RoMFAC & \textbf{0.92} & \textbf{63.40$\pm$1.99} & \textbf{316.93$\pm$14.78} & \textbf{3815.51$\pm$408.88} \\
		\midrule  
		\cellcolor[HTML]{FFFFFF}                     & MFAC        & 0.24          & 53.74$\pm$7.96          & 250.48$\pm$43.43          & 2356.47$\pm$369.75          \\ 
		\cellcolor[HTML]{FFFFFF}                     & SA-MFAC     & 0.36          & 57.20$\pm$5.78          & 276.58$\pm$29.36          & 2930.68$\pm$369.41          \\
		\cellcolor[HTML]{FFFFFF}                     & SA-MFAC$^3$ &0.32          & 56.90$\pm$5.49         & 282.87$\pm$28.02         & 3600.57$\pm$416.66     \\ 
		\cellcolor[HTML]{FFFFFF}                     & RoMFAC$^1$    & 0.54          & 60.58$\pm$4.67          & 299.88$\pm$24.71          & 3232.90$\pm$371.80          \\ 
		\multirow{-14}{*}{\cellcolor[HTML]{FFFFFF}16} & RoMFAC & \textbf{0.86} & \textbf{62.94$\pm$2.77} & \textbf{312.51$\pm$17.98} & \textbf{3724.85$\pm$394.78} \\ 
		\midrule  
		\cellcolor[HTML]{FFFFFF}                     & MFAC        & 0.00          & 42.28$\pm$7.96          & 185.45$\pm$42.56          & 1088.57$\pm$373.91          \\ 
		\cellcolor[HTML]{FFFFFF}                     & SA-MFAC     & 0.34          & 58.26$\pm$4.17          & 281.67$\pm$26.94          & 3015.21$\pm$450.06          \\
		\cellcolor[HTML]{FFFFFF}                     & SA-MFAC$^3$ &0.38           & 57.92$\pm$4.62         & 283.68$\pm$25.12         & 3555.26$\pm$381.79     \\ 
		\cellcolor[HTML]{FFFFFF}                     & RoMFAC$^1$    & 0.46          & 59.24$\pm$5.49          & 293.16$\pm$29.98          & 3171.95$\pm$448.23          \\ 
		\multirow{-14}{*}{\cellcolor[HTML]{FFFFFF}32} & RoMFAC & \textbf{0.88} & \textbf{63.40$\pm$1.69} & \textbf{315.15$\pm$16.26} & \textbf{3714.23$\pm$485.07} \\ 
		\midrule  
		\cellcolor[HTML]{FFFFFF}                     & MFAC        & 0.00          & 35.64$\pm$6.85          & 152.25$\pm$ 34.83         & 979.07$\pm$321.50           \\ 
		\cellcolor[HTML]{FFFFFF}                     & SA-MFAC     & 0.30          & 56.32$\pm$5.13          & 273.99$\pm$27.81          & 2866.90$\pm$437.40        \\ 
		\cellcolor[HTML]{FFFFFF}                     & SA-MFAC$^3$ &0.28          & 57.10$\pm$5.07         & 280.95$\pm$24.83        & 3598.86$\pm$460.70     \\
		\cellcolor[HTML]{FFFFFF}                     & RoMFAC$^1$    & 0.32          & 57.28$\pm$5.59          & 278.95$\pm$32.30          & 3165.30$\pm$421.10          \\ 
		\multirow{-14}{*}{\cellcolor[HTML]{FFFFFF}48} & RoMFAC & \textbf{0.88} & \textbf{63.04$\pm$2.88} & \textbf{308.64$\pm$20.75} & \textbf{3676.54$\pm$464.25} \\ 
		\midrule  
		\cellcolor[HTML]{FFFFFF}                     & MFAC        & 0.00          & 26.42$\pm$3.99          & 102.08$\pm$19.29          & 1000.10$\pm$388.11          \\ 
		\cellcolor[HTML]{FFFFFF}                     & SA-MFAC     & 0.24          & 55.58$\pm$5.00          & 265.56$\pm$22.69          & 2856.36$\pm$486.67          \\ 
		\cellcolor[HTML]{FFFFFF}                     & SA-MFAC$^3$ &0.24          & 56.20$\pm$5.13         & 275.29$\pm$25.09        & 3551.70$\pm$398.98     \\
		\cellcolor[HTML]{FFFFFF}                     & RoMFAC$^1$    & 0.26          & 56.40$\pm$5.72          & 275.40$\pm$31.16          & 3175.48$\pm$459.65          \\ 
		\multirow{-14}{*}{\cellcolor[HTML]{FFFFFF}64} & RoMFAC & \textbf{0.90} & \textbf{63.20$\pm$2.38}  & \textbf{308.48$\pm$18.51} & \textbf{3655.81$\pm$507.83} \\ 
		\bottomrule
	\end{tabular}}
	\caption{Performance comparisons among the proposed method and baselines. Bold scores represents the best performance.}
	\label{tab1}
\end{table}

\paragraph{Baselines and Ablations.} In experiments, we compare RoMFAC with MFAC and SA-MFAC, where MFAC does not apply any robust strategies and SA-MFAC, like most robust training, uses a fixed weight factor $\mu$ and increased perturbation bound $\epsilon$ in the original training technique. In addition, we apply our repetitive regularization technique to SA-MFAC, denoted as SA-MFAC$^3$, that is, $\mu$ remains constant, but $\epsilon$ changes repetitively. We also consider one variant of RoMFAC for ablation studies, namely the $\mu$ of RoMFAC just uses linear increase one time in the training process in order to show the effectiveness of our repetitive regularization, denoted as RoMFAC$^1$. 

\paragraph{Training.} In the battle scenario, we train five models in self-play. In the pursuit scenario, both predators and prey use the same algorithm during training process. In the two scenarios five models have almost the same hyper-parameters settings except the number of loops $c$. For MFAC, $\mu$ is always $0$ during the whole training process. For SA-MFAC and RoMFAC$^1$, we execute one loop for $\epsilon$ and $\mu$ (i.e., $c=1$), respectively. For SA-MFAC$^3$ and RoMFAC, we execute three loops for $\epsilon$ and $\mu$ (i.e., $c=3$), respectively. Settings of other hyper-parameters are put into Appendix E of our supplementary file.

\paragraph{Testing.} For the convenience of comparison, we use advantageous actor critic as the opponent agents' and prey's policies, and do not perturb their states. To evaluate the algorithm's robustness, we utilize a 10-step PGD with $\ell_{\infty}$ norm perturbation budget $\epsilon=0.075$ to create adversarial perturbations and execute 50 rounds of testing with maximum time steps of 400. 
\begin{figure}[t]
	\centering
	\subfigure[Battle.]{
		\begin{minipage}[t]{\linewidth}
			\centering
			\includegraphics[scale=0.48]{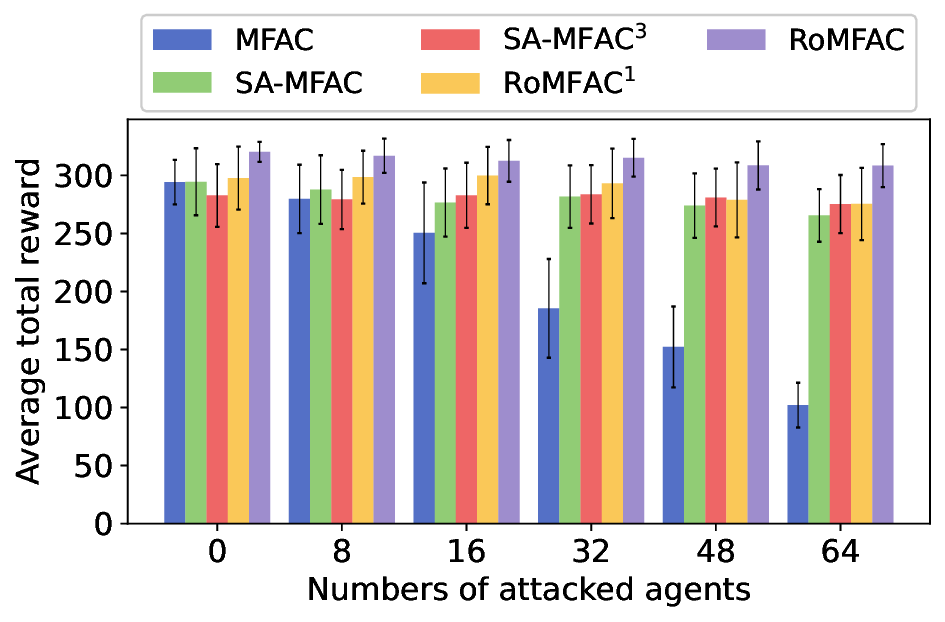}
		\end{minipage}%
	}%

	\subfigure[Pursuit.]{
		\begin{minipage}[t]{\linewidth}
			\centering
			\includegraphics[scale=0.48]{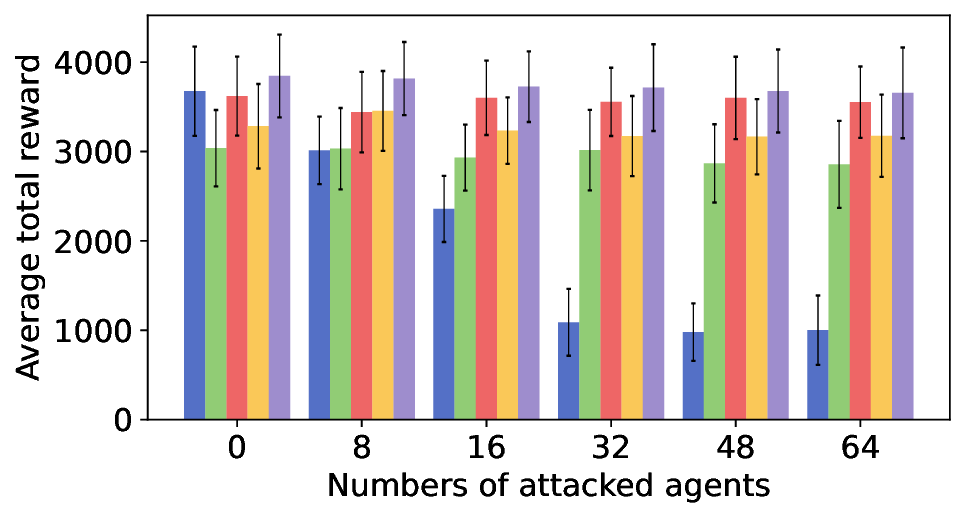}
		\end{minipage}%
		
	}%
	\caption{Average total reward.}
	\label{results}
\end{figure}

\subsubsection{Results and Discussions}  
As demonstrated in Figure \ref{figscenarios}, MFAC agents cannot collaborate normally when states are perturbed. In the battle scenario, the previously learned policies of collaboration that a group of agents collaboratively go forward and attack another group are destroyed. They begin attacking empty grids but are unable to accurately attack opponent agents. In the pursuit scenario, the initially learned coordinated siege policy is destroyed, their movements are scattered, and they are unable to accurately attack prey.

The experimental results are shown in Table \ref{tab1}. Figure \ref{results} presents the average total rewards of agents. It is seen that when the robust training is not carried out, the cooperative policies will be destroyed more seriously with more attacked agents. After a robust training, the performance of the model will slightly decrease as the number of attacked agents grows. In the battle scenario, a robust training will increase the performance not only on adversarial states but also on clean states. Our RoMFAC is the most effective. 
The winning rate and the number of opponent agents killed can also be used for performance evaluation. In the pursuit scenario, under clean states, the performance of SA-MFAC, SA-MFAC$^3$ and RoMFAC$^1$ approaches will decrease, whereas our RoMFAC method will improve the performance not only on adversarial states but also on clean states. In a word, the model trained by our RoMFAC has the better performance even in the environments without perturbations (i.e., the number of attacked agents is $0$).

Comparing RoMFAC$^1$ and RoMFAC, we can see the significance of our repetitive regularization. 
The average total rewards obtained by SA-MFAC$^3$ are slightly better than SA-MFAC in the battle scenario, but they are obviously good in the pursuit scenario. Therefore,  applying our repetitive regularization to SA-MFAC can also lead to a good result.

\section{Conclusion and Future Work}
In this paper, we propose a robust training framework for the state-of-the-art reinforcement learning method MFAC. In our framework, the action loss function and the repetitive regularization of it play an important role in improving the robustness of the trained model. Moreover, we present SASG to establish a theoretical foundation in multi-agent reinforcement learning with adversarial attacks and defenses, which shows that our proposed action loss function is convergent. Our work is inspired by SA-MDP \cite{zhang2020robust} that is a robust single-agent reinforcement learning with adversarial perturbations and a special case of SASG. In the future work, we intend to extend our method to other MARL approaches. 

\bibliographystyle{named}
\bibliography{ijcai22}
\end{document}